\documentclass[sigconf, 9pt]{acmart}

\usepackage{graphicx}
\usepackage{subcaption}
\usepackage{algorithmic}
\usepackage{algorithm}
\usepackage{multirow}
\usepackage{pifont}
\providecommand{\cmark}{\ding{51}}
\providecommand{\xmark}{\ding{55}}
\usepackage{dsfont}
\usepackage{booktabs}
\usepackage{tabularx}
\usepackage{array}
\usepackage[normalem]{ulem}
\useunder{\uline}{\ul}{}
\usepackage[table]{xcolor}
\usepackage{makecell}

\AtBeginDocument{%
  }

\acmYear{2027}\copyrightyear{2027}
\setcopyright{acmlicensed}

\begin{document}

\title{SenseAgent: An LLM Agent for Adaptive Cross-Domain IMU Sensing}

\author{Tianya Zhao}
\affiliation{%
  \institution{Florida International University}
  \city{Miami}
  \state{Florida}
  \country{USA}}
\email{tzhao010@fiu.edu}

\author{Chuan Liu}
\affiliation{%
  \institution{Florida International University}
  \city{Miami}
  \state{Florida}
  \country{USA}}
\email{cliu060@fiu.edu}

\author{Xuyu Wang}
\authornote{Corresponding Author}
\affiliation{%
  \institution{Florida International University}
  \city{Miami}
  \state{Florida}
  \country{USA}}
\email{xuywang@fiu.edu}

\begin{abstract}
Deep learning has improved inertial measurement unit (IMU) sensing for mobile and wearable applications. However, an IMU model trained in one domain often becomes unreliable when it is used with a new user, device, or body position. Existing methods usually treat this problem as a static model-design task: they pretrain a stronger representation, add data augmentation, or select one adaptation method before deployment. In practice, the target domain is only gradually observed, labels are scarce, and different domain shifts require different sensing actions.
This paper presents SenseAgent, an LLM-guided sensing agent for cross-domain IMU activity recognition. Instead of asking an LLM to classify raw IMU signals, SenseAgent uses the LLM as a runtime planner over sensing tools, source-domain experience memory, online target memory, and verifiers. The agent builds a label-free diagnosis report from the target stream and uses it to decide whether to keep raw inference or invoke specialized tools, including gravity-aware sensing, prototype transfer, and style normalization. Verifiers check source calibration, target-memory reliability, and no-harm criteria before accepting high-risk tool decisions. SenseAgent also supports scarce feedback without retraining the backbone or replacing the label-free route. This design converts cross-domain IMU sensing from a fixed inference pipeline into a closed-loop sensing process that diagnoses target shifts, selects suitable sensing actions, and rejects unsafe adaptations.
We evaluate SenseAgent across multiple IMU datasets and deployment shifts. Results show that its verified route selection improves cross-domain sensing, especially under harder placement and compound shifts, and further benefits from limited user feedback.
\end{abstract}





\maketitle

\section{Introduction}\label{sec:Intro}

Inertial measurement units (IMU) have become a foundational sensing modality for mobile and wearable computing. Smartphones, smartwatches, earbuds, AR/VR headsets, and specialized wearable devices routinely output data streams from accelerometers and gyroscopes, which can capture human motion, device interaction, mobility patterns, and physical activities. Recent deep learning techniques have substantially improved IMU-based sensing, enabling applications such as human activity recognition (HAR)~\cite{zhao2025membership, xu2021limu, jain2022collossl}, gesture recognition~\cite{zhang2021fine}, health monitoring~\cite{bo2022imu}, and human-computer interaction~\cite{hu2023combining}.

Despite these advances, IMU sensing still faces a fundamental cross-domain challenge. A deep learning model trained in one setting can perform poorly when deployed with a new user, device type, or device position, which creates an out-of-distribution (OOD) target domain relative to the training data. To address this issue, existing studies improve model generalization across domains using self-supervised pretraining~\cite{xu2021limu,hong2024crosshar}, data augmentation~\cite{li2020activitygan,zhou2024autoaughar}, few-shot learning~\cite{zhao2024cross,gong2019metasense}, and domain adaptation~\cite{chang2020systematic,zhou2022target}. 
However, these methods typically rely on a predefined or fixed adaptation pipeline. Once the model enters a target domain, the same procedure is applied regardless of the observed shift.

In real-world deployments, domain shifts are often heterogeneous and context-dependent. A performance drop may be caused by user-specific motion style, device rotation, body placement, or a combination of these factors. These shifts may affect different activities and users in different ways, making it difficult to determine a priori which adaptation strategy will be effective. Therefore, different target domains may require different actions. 
For example, gravity-aware reasoning may help when the dominant shift comes from orientation or placement, while prototype transfer may be more useful when activity patterns remain class-consistent but differ in amplitude. 
Applying the wrong action can cause negative transfer, especially when target labels are unavailable.

This observation suggests that cross-domain IMU adaptation should not be viewed only as an offline model optimization problem, but also as a deployment-time decision-making problem over sensing data, model diagnostics, and adaptation actions. An effective system should inspect the unlabeled target stream, summarize label-free evidence, choose an appropriate sensing action, verify whether the action is safe, and update its target memory as more data or scarce feedback becomes available. 

Recent advances in large language model (LLM) agents provide a new opportunity to rethink this process. Unlike standalone LLMs that primarily generate textual responses, LLM agents can interact with external tools, execute multi-step workflows, analyze intermediate results, and iteratively refine their decisions based on feedback. These capabilities make them a promising mechanism for automating complex sensing pipelines that require reasoning over data characteristics, model behavior, and adaptation outcomes. 
In the context of IMU sensing, an LLM agent could serve as a runtime planner that inspects target-domain evidence, selects diagnostic tools, configures adaptation modules, evaluates feedback, and refines the adaptation strategy for a specific deployment scenario.

However, directly applying general-purpose LLM agents to IMU sensing faces three significant challenges. 
First, the agent needs \emph{sensing-grounded observations}. IMU windows are continuous, noisy, and high-frequency time series whose semantics are not explicitly expressed in natural language~\cite{bian2026foundation,shen2025autoiot}. Raw waveforms do not directly expose the physical and statistical factors that matter for adaptation, such as gravity direction, motion intensity, temporal dynamics, or prototype consistency. Therefore, a useful agent needs compact sensing abstractions instead of raw IMU tokens. 
Second, the agent must make \emph{deployment-time decisions under label scarcity}. A target domain may differ from the source domains in user style, motion amplitude, temporal dynamics, device orientation, or body placement~\cite{caramaschi2023device,oishi2025wimusim}, and different shifts may require different adaptation actions. 
However, during deployment, the system observes mostly unlabeled target windows and may receive only limited user feedback. Thus, the agent cannot select an action by directly measuring target-domain accuracy; it must infer the appropriate action from source-domain experience and label-free evidence in the target data.
Third, the agent must avoid \emph{negative transfer}. Different adaptation tools are useful for different shifts, and an adaptation that improves one target domain may degrade another~\cite{pan2009survey,wang2019characterizing}. For example, orientation-aware correction can mitigate device-rotation or body-placement shifts, but it may suppress useful motion variations when the dominant shift instead comes from user-specific motion style or device heterogeneity.

To address these challenges, we propose \textit{SenseAgent}, an LLM-agent framework for adaptive cross-domain sensing. SenseAgent uses the LLM as a structured runtime planner rather than a raw-signal classifier. It does not directly predict labels from IMU windows. Instead, it reasons over structured evidence produced by local sensing modules and coordinates adaptation actions. 
To provide sensing-grounded observations, SenseAgent converts the target stream into compact diagnostic reports that summarize motion evidence, gravity consistency, prototype agreement, model confidence, tool disagreement, and optional user feedback.
To support deployment-time decisions under label scarcity, SenseAgent builds source experience memory offline and updates target-domain memory online, allowing the planner to reason over domain-level context rather than isolated windows or unavailable target labels. 
To reduce negative transfer, SenseAgent restricts the action space to auditable sensing actions, such as tool selection, conservative fusion, reliable memory update, and memory freezing. A verifier checks each planned tool invocation and memory update using label-free stability signals and optional user feedback before changing the prediction rule.
This paper makes the following contributions:
\begin{itemize}
    \item To the best of our knowledge, this is the first work to systematically investigate the LLM agent to improve cross-domain IMU sensing performance. We formulate cross-domain IMU adaptation as an agentic sensing problem, where the system makes sequential decisions over observations, memory, tool selection, feedback, and verification.
    \item We propose SenseAgent, an LLM-guided sensing agent that integrates source experience memory, online target memory, motion-aware diagnosis, and verifier-controlled action selection with a set of sensing tools, including dynamic-anchored gravity reasoning, raw and gravity-view prototype experts, style normalization, and feedback-based propagation.
    \item We evaluate SenseAgent on multiple IMU datasets and OOD settings covering user, sensing-condition, and compound shifts. The evaluation studies both zero-feedback and scarce-feedback adaptation, and analyzes how planner decisions, verifier overrides, and stronger LLM planners affect final sensing performance.
\end{itemize}

\section{Background and Motivation}\label{sec:motivation}

\subsection{Background: LLM Agents}

LLMs have shown strong capabilities in natural language understanding, reasoning, and code generation, but a standalone LLM typically acts as a passive text generator that produces one response from a given prompt. In contrast, an LLM agent uses the LLM as a reasoning and planning module within an interactive system~\cite{luo2025large}. Such an agent can decompose tasks, invoke external tools, observe intermediate results, and update later decisions based on feedback.


This paper focuses on three agent functionalities that are most relevant to sensing-system adaptation. 
First, \emph{reasoning and planning} enables an agent to decide which actions to take and in what order. ReAct demonstrates this by interleaving reasoning with task-specific actions~\cite{yao2022react}. 
Second, \emph{tool use} allows the agent to access external capabilities such as retrieval, computation, and program execution. Toolformer shows that language models can learn when and how to call external APIs~\cite{schick2023toolformer}. 
Third, \emph{feedback and memory} allow an agent to store observations, reflect on previous outcomes, and revise later decisions, as illustrated by Reflexion~\cite{shinn2023reflexion}.


These functionalities make LLM agents attractive for system tasks that require multi-step decision-making rather than one-shot generation. 
Recent Artificial Internet of Things (AIoT) work has used LLM agents to translate natural-language requirements into programs and refine them with execution feedback~\cite{shen2025autoiot}. However, IMU sensing poses different challenges: raw sensor streams are noisy, non-linguistic, and environment-dependent, while adaptation decisions must be grounded in physical signals, data distributions, and model behavior. Therefore, applying LLM agents to IMU sensing requires domain-specific observations, tools, and verification mechanisms, rather than directly prompting a general-purpose LLM with raw sensor data.

\subsection{Preliminary Experiments}

\begin{figure}[t]
    \centering
    \begin{subfigure}{0.44\columnwidth}
        \centering
        \includegraphics[width=\linewidth]{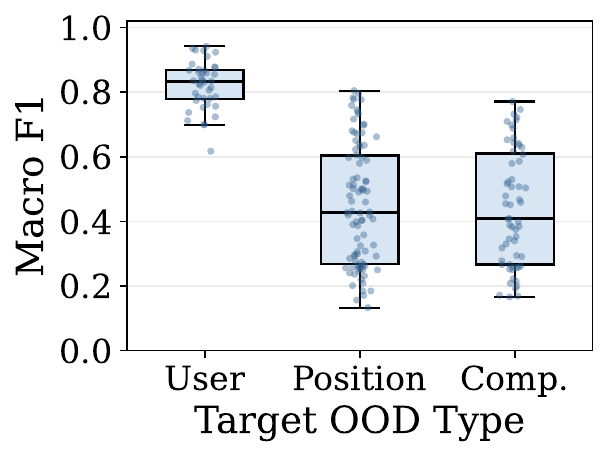}
        \caption{Difficulty distribution}
        \label{fig:pre_1_a}
    \end{subfigure}
    \hfill
 \begin{subfigure}{0.545\columnwidth}
        \centering
        \includegraphics[width=\linewidth]{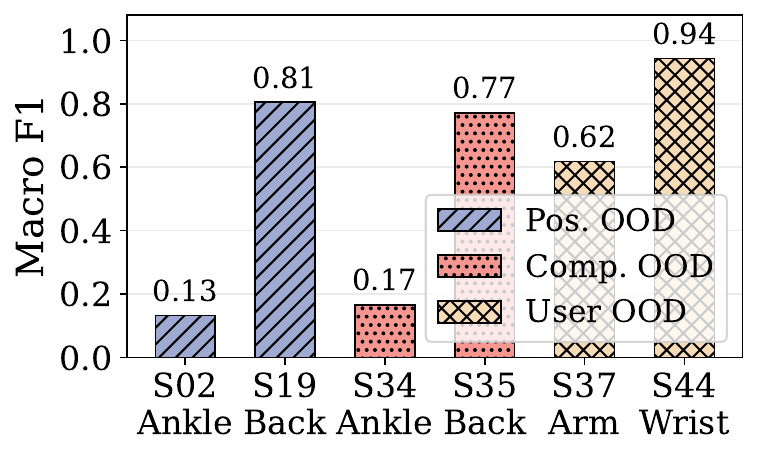}
        \caption{Representative target domains}
        \label{fig:pre_1_b}
    \end{subfigure}
    \caption{OOD difficulty is not uniform across target domains. (a) Raw PatchTST macro-F1 on TNDA varies substantially across users, positions, and compound OOD settings. (b) Representative target domains show that the same OOD type can include both easy and hard cases.}
    \label{fig:pre_1}
\end{figure}

\begin{figure}[t]
    \centering
    \begin{subfigure}{0.5\columnwidth}
        \centering
        \includegraphics[width=\linewidth]{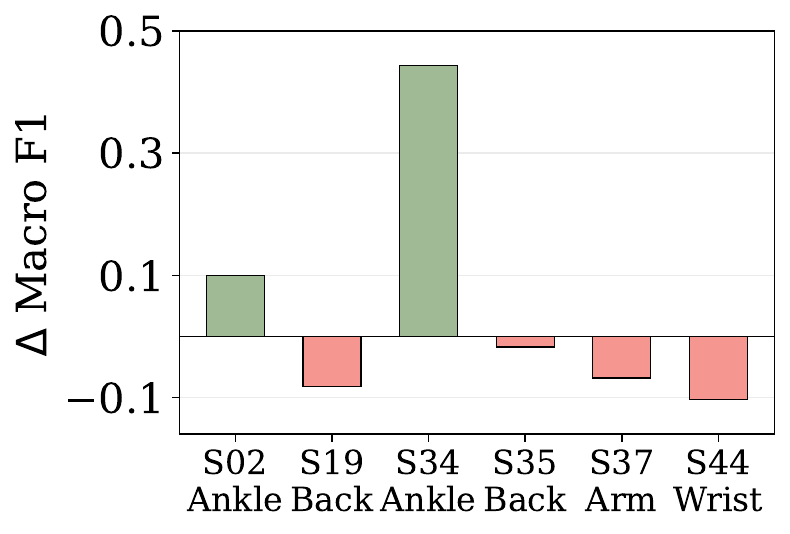}
        \caption{1-shot adaptation}
        \label{fig:pre_2_a}
    \end{subfigure}
    \hfill
 \begin{subfigure}{0.47\columnwidth}
        \centering
        \includegraphics[width=\linewidth]{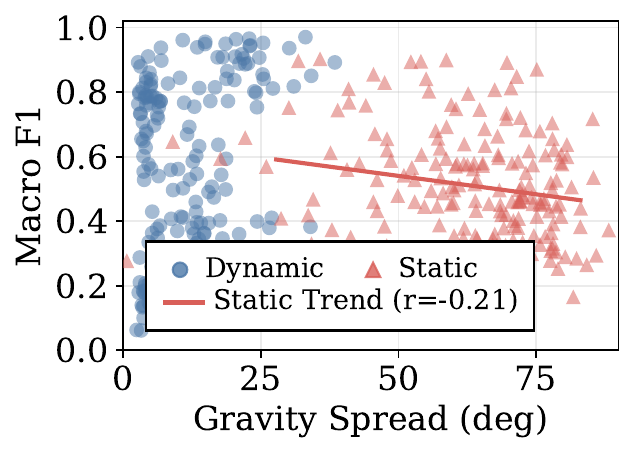}
        \caption{Groupwise stability}
        \label{fig:pre_2_b}
    \end{subfigure}
    \caption{Adaptation and physical cues are target-dependent on TNDA.
    (a) A 1-shot prototypical baseline can either improve or degrade target domains.
    (b) Dynamic windows provide a stable gravity reference, whereas static windows show larger dispersion. Static dispersion has a weak negative association with raw static-class F1.}
    \label{fig:pre_2}
\end{figure}

We first examine whether cross-domain IMU deployment can be handled by a fixed adaptation recipe. 
In practice, a sensing model may face different target domains over time, including new users, new sensor positions, or their combinations, where the same adaptation strategy may help one domain but hurt another.
We study this variability on TNDA~\cite{tnda2021dataset}, which provides diverse user and position shifts, by training PatchTST~\cite{nie2022time} on source domains and evaluating each held-out target domain independently.


\textbf{Observation 1: OOD difficulty is domain-specific.}
Fig.~\ref{fig:pre_1} shows that cross-domain performance varies widely even within the same OOD category. 
Some user-shift domains remain close to the source distribution, whereas some position and compound domains suffer large drops, indicating that placement-related changes often create the hardest deployment cases. 
The representative examples in Fig.~\ref{fig:pre_1_b} show the same pattern at deployment granularity: an ankle-mounted target domain can be challenging, while a back-mounted target domain under the same split can be much easier. 
Therefore, the OOD category alone is not sufficient to determine how the system should respond.

\textbf{Observation 2: Adaptation is not uniformly beneficial.}
Fig.~\ref{fig:pre_2_a} evaluates a simple 1-shot prototypical adaptation baseline on representative TNDA domains. The effect is mixed: it substantially improves hard ankle cases, such as Subject34-LeftAnkle, but degrades several easier domains, such as Subject19-Back and Subject44-RightWrist. This behavior is expected: a few target examples can correct a shifted local neighborhood, but they can also perturb a source model that is already reliable. Thus, scarce target feedback or few-shot prototypes should not be applied blindly.

\textbf{Observation 3: Gravity evidence differs between static and dynamic activities.}
Fig.~\ref{fig:pre_2_b} shows that dynamic windows have a much more concentrated gravity spread than static windows, making them a stable label-free proxy for estimating the target gravity reference. 
Static windows show larger dispersion, suggesting stronger exposure to gravity geometry. The negative but weak correlation with raw static-class F1 ($r=-0.21$) indicates that gravity is useful but insufficient: it can reveal some placement-induced static ambiguity, especially for lying-like postures, but upright postures may remain close to the dynamic gravity direction. 
Therefore, gravity should be treated as a selective diagnostic cue for static activities, while dynamic activities require motion-pattern evidence beyond gravity.

\subsection{Motivation}

The preliminary study identifies three empirical facts that motivate SenseAgent. First, OOD difficulty is target-specific: domains under the same shift category can exhibit very different performance. Second, adaptation is not uniformly safe: a prototype or feedback-based action may improve a hard target domain but degrade an easy one. Third, gravity provides a useful label-free physical cue, but its role is activity-dependent: it helps diagnose some static placement ambiguity, while dynamic activities may require more motion-pattern evidence beyond gravity.

These facts show that cross-domain IMU adaptation cannot be handled by a single static model or a preselected adaptation rule. In realistic deployment, the system does not know whether the dominant shift comes from user style, sensor placement, gravity misalignment, or their combination. It also cannot directly measure target accuracy because labels are unavailable or scarce. The system must therefore make a sequence of deployment-time decisions: inspect label-free target evidence, choose a sensing action, verify whether the action is safe, and update memory as more target data or feedback becomes available.

SenseAgent follows this agentic design. The LLM does not classify IMU windows directly. Instead, it maintains source and target memory, reasons over structured sensing diagnostics, and plans over a constrained set of sensing actions. Local tools execute numerical IMU operations, including gravity diagnostics, prototype matching, feedback integration, and conservative fusion. Verifier modules check whether tool outputs and memory updates are supported before they affect final predictions. This allows SenseAgent to adapt to heterogeneous target domains while grounding every decision in measurable IMU evidence.

\section{Problem Formulation}\label{sec:problem}

We consider cross-domain IMU activity recognition, where labeled data are available from one or more source domains and the target domain is observed only at deployment time. Let $\mathcal{Y}=\{1,\ldots,C\}$ denote the activity label space. We denote the collection of $S$ labeled source domains as:
\begin{equation}
    \mathcal{D}_S = \{\mathcal{D}_s\}_{s=1}^{S}, \quad
    \mathcal{D}_s = \{(x_i^s,y_i^s)\}_{i=1}^{n_s},
\end{equation}
where $\mathcal{D}_s$ is the $s$-th source domain, $n_s$ is the number of labeled windows in that domain, and $y_i^s\in\mathcal{Y}$ is the activity label. Each IMU window is a length-$L$ six-axis time series:
\begin{equation}
    x_i^s = \{a_i^s(\tau), g_i^s(\tau)\}_{\tau=1}^{L}, \quad
    a_i^s(\tau), g_i^s(\tau) \in \mathbb{R}^{3},
\end{equation}
where $a_i^s(\tau)$ and $g_i^s(\tau)$ denote the three-axis accelerometer and gyroscope readings at time step $\tau$, respectively. Depending on the dataset split, a domain may correspond to a user, device, body position, sensor placement, or other deployment condition.

At deployment time, the system receives a target stream:
\begin{equation}
    \mathcal{X}_{\mathrm{tar}} = \{x_k^t\}_{k=1}^{\infty}, \quad
    x_k^t = \{a_k^t(\tau), g_k^t(\tau)\}_{\tau=1}^{L}.
\end{equation}
The target-domain identity, such as the user, device, or body position, is not assumed to be known. The activity labels of target windows are also unavailable during deployment, except for optional user feedback.
We denote the feedback set as:
\begin{equation}
    \mathcal{F}_{\mathrm{tar}} = \{(x_j^t,y_j^t): j \in \Omega\}, \quad
    \Omega \subset \mathbb{N}, \quad |\Omega|=N,
\end{equation}
where $\Omega$ contains the target windows explicitly labeled by the user and $N$ is the feedback budget. We consider both zero-feedback adaptation with $N=0$ and scarce-feedback adaptation with $N>0$.

The goal is to predict $\hat{y}_k^t\in\mathcal{Y}$ for each incoming target window $x_k^t$ under two constraints. First, target labels are unavailable except for explicitly provided feedback. Second, target-domain identities are not used for adaptation or prediction. Therefore, the system must adapt to an unknown target domain using labeled source data, an unlabeled target stream, and optional scarce feedback, without using hidden target labels or domain identifiers.

\section{SenseAgent Design}\label{sec:design}

\subsection{System Overview}

SenseAgent is a verifier-guarded LLM sensing agent for cross-domain IMU activity recognition. The LLM is used as a structured runtime planner rather than an IMU classifier. It does not process raw IMU signals or directly predict activity labels. Instead, local sensing modules convert the target stream into compact evidence, the planner proposes a structured sensing action, and verifiers decide whether the action is safe to execute.

We use four terms consistently throughout the design. \emph{Source experience memory} contains offline assets built only from labeled source domains, including the frozen backbone, source statistics, prototype banks, and source-calibrated reliability summaries. \emph{Online target memory} is a verified cache of unlabeled target windows and optional feedback anchors; it provides target-side context without retraining the backbone. A \emph{tool} is a local computation module that produces evidence, candidate logits, or reliability scores. A \emph{route} is an executable prediction path accepted by the verifier, which combines one or more tools and outputs final logits.

\begin{figure*}
    \centering
    \includegraphics[width=0.9\linewidth]{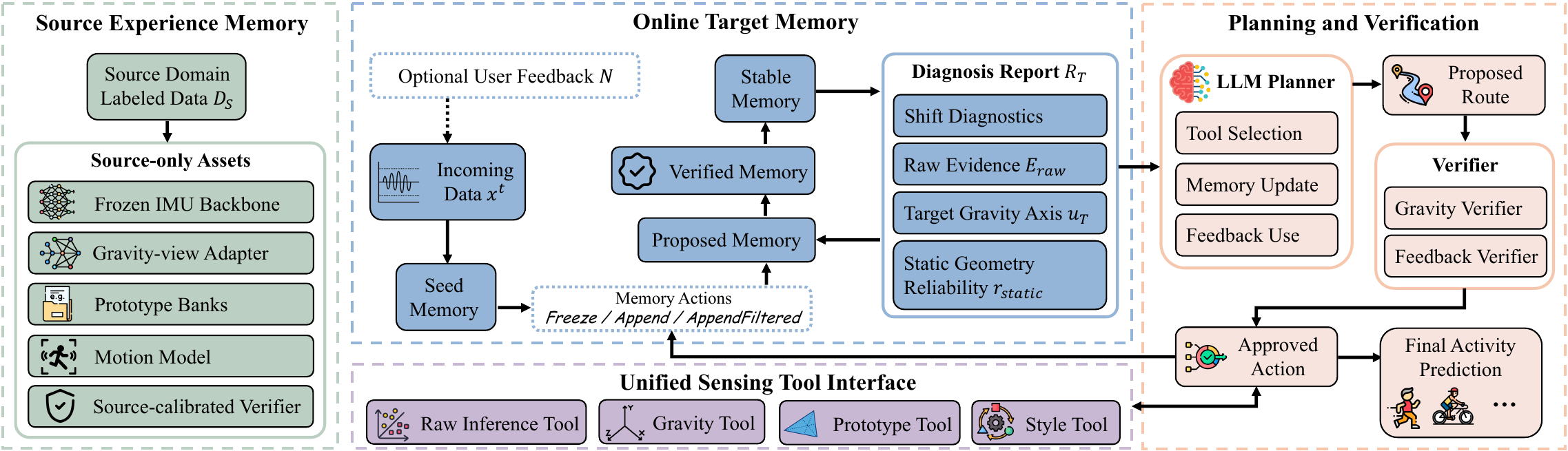}
    \caption{Overview of SenseAgent.}
    \label{fig:overview}
\end{figure*}

Fig.~\ref{fig:overview} shows the overall workflow. 
Offline, SenseAgent builds source-only assets. Online, each target buffer is encoded by the frozen backbone, summarized with the motion model and target memory, and converted into a diagnosis report. 
The LLM planner then proposes a prediction route, a memory action, and optional feedback use. The verifier checks source support, target-memory stability, and no-harm conditions before any route or memory update affects predictions.

\subsection{Unified Sensing Tool Interface}

SenseAgent exposes four complementary sensing tools to the planner. 
The raw inference tool provides the conservative source-model logits. The gravity tool constructs gravity-aligned views and gravity reliability diagnostics for orientation-related shifts. The prototype tool provides class-level dynamic evidence in either the raw embedding space or the gravity-view embedding space. The style tool estimates channel-statistics shifts and produces normalized candidate logits. 
These tools do not directly determine final predictions; they only provide the evidence and candidate outputs used by the planner, the verifier, and the verified routes described next.

\textbf{Raw inference tool.}
The raw inference tool is the default prediction rule. It applies the frozen IMU model stored in the source memory to the original six-axis input. Given a target window $x_i^t$, the raw logits $\ell_i^0$ and probabilities $P_i^0$ are
\begin{equation}
    \ell_i^0 = h_\theta(f_\theta(x_i^t)), \quad
    P_i^0 = \operatorname{softmax}(\ell_i^0).
\end{equation}
where $f_\theta$ denotes the encoder and $h_\theta$ denotes the classification head.

\textbf{Gravity tool.}
The gravity tool provides an orientation-aware view for domains where device rotation or body placement may distort raw channels. 
Given an estimated gravity axis $u$, the tool converts a six-axis IMU window into gravity-aligned channels:
\begin{equation}
\left\{
\begin{aligned}
    a_v(\tau) &= a(\tau)^\top u, 
    & g_v(\tau) &= g(\tau)^\top u, \\
    a_h(\tau) &= \left\|a(\tau)-a_v(\tau)u\right\|_2,
    & g_h(\tau) &= \left\|g(\tau)-g_v(\tau)u\right\|_2, \\
    a_n(\tau) &= \left\|a(\tau)\right\|_2,
    & g_n(\tau) &= \left\|g(\tau)\right\|_2 .
\end{aligned}
\right.
\label{eq:gravity_features}
\end{equation}
The resulting view supports two forms of evidence. For low-motion windows, it provides geometric evidence for separating lying from upright postures. For dynamic windows, it provides gravity-view embeddings for prototype matching.

\textbf{Prototype tool.}
The prototype tool provides class-level dynamic evidence. SenseAgent includes both raw-view and gravity-view prototype tools. They share the same retrieval and refinement interface but operate in different embedding spaces. Given dynamic support embeddings $z_i$ from target memory and retrieved source prototypes $p_c^{\mathrm{src}}$, the tool computes soft prototype assignments:
\begin{equation}
    Q_{ic}=\operatorname{softmax}_c\left(-\|\operatorname{norm}(z_i)-p_c^{\mathrm{src}}\|_2^2\right).
\end{equation}
The $\ell_2$ normalization makes prototype matching depend on embedding direction rather than domain-specific magnitude. The assignment matrix $Q$ summarizes how the target dynamic support aligns with source class structure. The tool then forms target-supported class means: $\hat{p}_c=\operatorname{norm}(\sum_i Q_{ic}\operatorname{norm}(z_i)/(\sum_i Q_{ic}+\epsilon))$ and reports candidate prototype logits together with reliability diagnostics.
The tool itself does not decide whether these target-supported means should change the final prediction. That decision is deferred to the source-calibrated verifier and the verified sensing route.



\textbf{Style tool.}
The style tool compensates for channel-level distribution shifts by mapping target channel statistics to source statistics:
\begin{equation}
    x^{\mathrm{style}}
    =
    (x-\mu_T)\frac{\sigma_S}{\sigma_T+\epsilon}+\mu_S,
\end{equation}
where \(\mu_T\) and \(\sigma_T\) are the per-channel mean and standard deviation estimated from target memory, while \(\mu_S\) and \(\sigma_S\) are the corresponding source statistics.
The tool is considered reliable only when channel-style shift is evident, the transformed prediction remains consistent with raw inference, and uncertainty decreases.


\subsection{Source Experience Memory}\label{sec:source_memory}

The source experience memory stores the source-only assets required by the sensing tools. It is constructed before deployment and never uses target-domain labels or target-domain identities.

\textbf{Frozen IMU backbone.}
SenseAgent first trains the backbone used by the raw inference tool and embedding-based tools on labeled source domains. Let $f_\theta$ denote the encoder and $h_\theta$ denote the classification head. The training objective is
\begin{equation}
    \min_\theta
    \sum_{s=1}^{S}\sum_{i=1}^{n_s}
    l_{CE}\left(h_\theta(f_\theta(x_i^s)),y_i^s\right),
\end{equation}
where $l_{CE}$ is the cross-entropy loss. 

\textbf{Gravity-view adapter.}
To support the gravity-view tool, we estimate a gravity axis $u_d$ from dynamic windows:
\begin{equation}
    u_d = \frac{\sum_{i\in\mathcal{I}_d^{\mathrm{dyn}}}\bar{a}_i}{\|\sum_{i\in\mathcal{I}_d^{\mathrm{dyn}}}\bar{a}_i\|_2+\epsilon}, \quad
    \bar{a}_i = \frac{\operatorname{median}_{\tau} a_i(\tau)}{\|\operatorname{median}_{\tau} a_i(\tau)\|_2+\epsilon},
    \label{eq:acc_direction}
\end{equation}
where \(a_i(\tau)\in\mathbb{R}^3\) denotes the acceleration vector and $\bar{a}_i$ is the normalized acceleration direction. The temporal median provides a robust gravity proxy: it suppresses short transient accelerations while preserving the dominant acceleration direction.
Then, IMU data are converted using Eq.~\ref{eq:gravity_features}. 
The default gravity-view encoder uses a lightweight low-rank adaptation (LoRA) trained on these gravity-view windows, while the base IMU backbone remains frozen. For an adapted linear layer with frozen weight \(W_\ell\), LoRA computes:
\begin{equation}
    y =
    W_\ell \cdot \xi + b_\ell
    + \frac{\alpha}{r}B_\ell \cdot A_\ell \cdot \xi,
\end{equation}
where \(\xi\) is the layer input, \(b_\ell\) is the bias, and \(A_\ell\in\mathbb{R}^{r\times d_{\mathrm{in}}}\) and \(B_\ell\in\mathbb{R}^{d_{\mathrm{out}}\times r}\) are the only trainable parameters for that layer. In this paper, we set \(r=4\), \(\alpha=8\).


\textbf{Prototype banks.}
With the raw and gravity-view encoders fixed, SenseAgent materializes prototype banks for dynamic activities. The system stores three banks: a raw-view bank from the frozen IMU backbone, a LoRA gravity-view bank from the adapted gravity encoder, and a base gravity-view bank that directly encodes gravity-view channels with the frozen backbone. 
For view $v$, source domain $d$, and dynamic class $c$, the stored prototype is
\begin{equation}
    p_{d,c}^{v}
    =
    \operatorname{norm}\left(
    \frac{1}{|\mathcal{I}_{d,c}|}
    \sum_{i\in\mathcal{I}_{d,c}} f^{v}(x_i)
    \right),
\end{equation}
where $f^v$ is the corresponding frozen encoder, $\mathcal{I}_{d,c}$ contains data from domain $d$ and class $c$, and $\operatorname{norm}(\cdot)$ denotes $\ell^2$ normalization. 

\textbf{Motion model.}
SenseAgent trains a source-only motion model to separate static and dynamic windows. For a window $x$, it computes a two-dimensional descriptor:
\begin{equation}
    \phi_m(x)
    =
    \left[
    \operatorname{std}_{\tau}\|a(\tau)\|_2,
    \sqrt{\operatorname{mean}_{\tau}\|g(\tau)\|_2^2}
    \right],
\end{equation}
where the two components capture acceleration variation and rotational energy. Because motion magnitudes vary across domains, we normalize this descriptor using source static windows as a baseline. Let $c_m$ and $s_m$ denote the median and interquartile range of $\phi_m(x)$ over source static windows. The dynamic probability is estimated by a logistic classifier:
\begin{equation}
    p_{\mathrm{dyn}}(x)=\sigma\left(w^\top ((\phi_m(x)-c_m)/(s_m+\epsilon))+b\right),
\end{equation}
where $\sigma(\cdot)$ is the sigmoid function, $w$ and $b$ are learned from source-domain labels, and $\epsilon$ is a small constant for numerical stability. The resulting score $p_{\mathrm{dyn}}(x)$ provides a source-normalized motion cue for diagnosis and tool selection during online deployment.

\textbf{Source-calibrated verifier.}
SenseAgent calibrates reliability models using leave-one-domain-out (LODO) source episodes. In each episode, one source domain is removed from prototype construction and treated as a pseudo-target stream. 
The system computes the same label-free signals that will be available online, while the pseudo-target labels are used only offline to determine whether a candidate action improves dynamic macro-F1 over the base logits. This produces supervision for reliability estimation rather than a target-domain adaptation rule.

For a candidate prototype action, the verifier uses the dynamic support set $\mathcal{I}_{\mathrm{dyn}} = \{i\in\mathcal{M}: p_{\mathrm{dyn}}(x_i)\ge 0.5\}$ with size $n=|\mathcal{I}_{\mathrm{dyn}}|$, and the soft assignment matrix $Q$ computed by the prototype tool. Each row of $Q$ represents how a dynamic support window is assigned to the candidate dynamic classes. The verifier summarizes this evidence as:
\begin{equation}
    q = \left[\log(1+n), H(Q), \min_c\sum_i Q_{ic}, \Delta_{\mathrm{ret}}, \mathds{1}_{\mathrm{bank}}\right],
\end{equation}
where $\log(1+n)$ measures the amount of dynamic data. The entropy term $H(Q)=\frac{1}{n}\sum_{i\in\mathcal{I}_{\mathrm{dyn}}}\left(-\sum_c Q_{ic}\log(Q_{ic}+\epsilon)\right)$ measures the uncertainty of the soft assignments. A low value indicates that support windows are assigned to prototypes with high confidence, whereas a high value indicates ambiguous prototype evidence. 
The term $\min_c\sum_i Q_{ic}$ measures whether the support is distributed across dynamic classes rather than collapsing to one class. 
The retrieval margin \(\Delta_{\mathrm{ret}}\) is the gap between the best and second-best source-domain matches for the current target support. A larger margin indicates that the target support can be matched to a source domain more confidently.
The indicator $\mathds{1}_{\mathrm{bank}}$ specifies whether a retrieved source-bank prior is used. For the global source prior, $\Delta_{\mathrm{ret}}=0$ and $\mathds{1}_{\mathrm{bank}}=0$. If $n=0$, all assignment-based features and the retrieval margin are set to zero.


Then, the verifier outputs a reliability score:
\begin{equation}
    r_v=
    \sigma\left(
    \eta^\top \frac{q-\mu_q}{\sigma_q+\epsilon}+b_q
    \right).
\end{equation}
The parameters are fitted from LODO rows $(q_j,y_j)$, where $y_j=1$ if the candidate prototype action improves dynamic macro-F1 over the corresponding base logits on the held-out source domain. Specifically, $\mu_q$ and $\sigma_q$ are the empirical mean and standard deviation of the LODO feature vectors, while $\eta$ and $b_q$ are the coefficient vector and intercept of a class-balanced logistic regression trained on the standardized features $(q_j-\mu_q)/(\sigma_q+\epsilon)$. 
The LODO summaries also store source-side support scores for each candidate tool, which are later combined with target diagnostics during verification.

\subsection{Online Target Memory}

At deployment, SenseAgent cannot access target labels but still needs target-side context to decide which tools are safe to use. Therefore, SenseAgent maintains an online target memory $\mathcal{M}_T$, a controlled cache of recent target windows used only for tool evidence and local statistics. 

\textbf{Memory states and updates.}
SenseAgent keeps three memory states: a seed memory with the initial target context, a stable memory approved by the verifier, and a proposed memory generated by the planner.
The planner can \textsc{Freeze} the memory, \textsc{Append} the current buffer, or \textsc{AppendFiltered} high-quality buffered windows while retaining feedback anchors. 
For filtered appending, each buffered window is ranked by a quality score $\rho_i=0.45C_i+0.35A_i+0.20K_i,$ where $C_i=\max_c P_0(c|x_i)$ is raw prediction confidence, $A_i$ is the largest vote fraction among local tool predictions, and $K_i=2|p_{\mathrm{dyn}}(x_i)-0.5|$ is motion clarity.

\textbf{Diagnosis report.}
The diagnosis report $R_T$ summarizes the current stable memory for the planner and verifier. For the raw model, the report includes memory-averaged normalized entropy $H_0$
and top-class margin $M_0$, which is the average probability gap between the largest and second-largest raw classes. 
These signals form raw positive evidence:
\begin{equation}
    E_{\mathrm{raw}}=0.5(1-H_0)+0.5M_0.
    \label{eq:raw_evidence}
\end{equation}
The report also records label-free shift diagnostics against raw inference. For example, prediction disagreement supports style reliability and groupwise arbitration, while the Jensen-Shannon divergence between raw and gravity logits contributes to the shift score used for action selection.


For gravity-view candidates, $R_T$ estimates a target gravity axis from verified memory. Given dynamic support windows
\(\mathcal{I}_{\mathrm{dyn}}=\{i\in\mathcal{M}_T:p_{\mathrm{dyn}}(x_i^t)\ge 0.5\}\), the target gravity axis is
\begin{equation}
    u_T =
    \frac{
    \sum_{i\in\mathcal{I}_{\mathrm{dyn}}}
    p_{\mathrm{dyn}}(x_i^t)\bar{a}_i
    }{\left\|\sum_{i\in\mathcal{I}_{\mathrm{dyn}}}
    p_{\mathrm{dyn}}(x_i^t)\bar{a}_i
    \right\|_2+\epsilon},
\end{equation}
where $\bar{a}_i$ is the acceleration direction defined in Eq.~\ref{eq:acc_direction}.

For low-motion windows \(\mathcal{I}_{\mathrm{stat}}=\{i\in\mathcal{M}_T:p_{\mathrm{dyn}}(x_i^t)<0.5\}\), the report computes signed angular evidence \(e_i^{\mathrm{geo}}\) for
lying-versus-upright geometry. The static geometry reliability is defined as:
\begin{equation}
    r_{\mathrm{static}}
    =
    c_{\mathrm{axis}}\cdot q_{\mathrm{clu}}\cdot
    \frac{1}{|\mathcal{I}_{\mathrm{stat}}|}
    \sum_{i\in\mathcal{I}_{\mathrm{stat}}}
    |e_i^{\mathrm{geo}}|,
\end{equation}
where \(c_{\mathrm{axis}}\) measures the concentration of dynamic-window gravity directions around \(u_T\), and \(q_{\mathrm{clu}}\) measures the quality of the two-cluster structure among low-motion gravity angles. SenseAgent sets \(r_{\mathrm{static}}=0\) when \(\mathcal{I}_{\mathrm{stat}}\) is empty. 

\textbf{Memory check.}
For each memory proposal, SenseAgent recomputes the diagnosis report $R_T$ and accepts the proposal only if it passes a no-harm memory check. 
A proposed memory is accepted only if it preserves the best label-free action value and does not weaken the verifier's reliability signals. When feedback anchors are available, it must also preserve feedback utility. Otherwise, SenseAgent keeps the stable memory.

\subsection{Agent Planning and Verification}

SenseAgent follows a report-action-verification pattern. At each decision point, local sensing modules first compute a diagnosis report from the verified target memory. The LLM planner then reads this report and outputs a constrained structured action:
\begin{equation}
    \{\texttt{route},\texttt{feedback\_use},\texttt{memory\_update},\texttt{rationale}\}.
\end{equation}
The planner does not freely call tools, execute code, or explore the environment. It serves as a sensing-specific arbitration layer over a fixed action space. Its role is to jointly choose the prediction route, memory update, and feedback use from heterogeneous evidence that is difficult to collapse into a single static rule. Deterministic local modules compute all logits, statistics, memory updates, and no-harm checks. 
The planner also returns a short textual rationale for auditing and debugging, but this rationale is not used as a numerical signal by the verifier or any prediction route.
Appendix~\ref{appx:prompt} provides the planner prompt template.


\textbf{Action evidence.}
During deployment, the planner reads the diagnosis report $R_T$ and converts it into evidence scores.
These scores summarize how strongly the current target memory supports each candidate route or tool module.
Raw evidence is given by $E_{\mathrm{raw}}$ in Eq.~\ref{eq:raw_evidence}, where high confidence and large margins favor keeping the frozen classifier.

For specialized tools and routes, evidence is derived from the diagnosis report most relevant to the corresponding shift. 
Gravity evidence combines static posture geometry and the source-calibrated reliability of gravity-view prototype matching.
Prototype evidence combines prototype reliability with the source-retrieval margin, which indicates whether the target support matches a source domain unambiguously. 
Style evidence favors normalization only when channel-style shift is evident, raw and style predictions remain consistent, and uncertainty is reduced. 
Groupwise evidence favors splitting static and dynamic windows only when the group utilities are positive.
Shift diagnostics further modulate these action values. Prediction disagreement and Jensen--Shannon divergence indicate when raw inference departs from alternative sensing views, but they do not directly determine activity labels.
The final prediction is produced only after the planner proposes a route and the verifier accepts or modifies it.

\textbf{Gravity verifier.} 
Gravity-based correction can mitigate placement shifts, but it can also cause negative adaptation when the target gravity reference is unreliable. Therefore, SenseAgent verifies gravity-aware actions in two stages.

The first stage is a route-level check that decides whether the gravity-aware route should be used. Let $u_A$ be the planner-selected route. 
If $u_A=\texttt{gravity-aware\_route}$, the verifier checks:
\begin{equation}
    r_{\mathrm{static}} < \tau_s \quad \wedge \quad E_{\mathrm{raw}} \ge \tau_r.
\end{equation}
This condition means that static-gravity evidence is weak while raw inference remains confident. When it holds, SenseAgent replaces the gravity action with \texttt{safe\_groupwise}; otherwise, it accepts the gravity route.
We use \(\tau_s=0.45\) and \(\tau_r=0.75\) in this paper.

The second stage is an implementation check within the accepted gravity-aware route. It selects between the LoRA gravity-view encoder and the base no-adapter gravity-view encoder using source support and target diagnostics. 
Overall, the route-level check decides whether the gravity route is safe to use, and the implementation check selects the better-supported gravity implementation.

\textbf{Feedback policy.}
The feedback budget $N$ specifies how many labeled target anchors are available during deployment. 
When $N=0$, SenseAgent uses the verified label-free route directly. When $N>0$, the labeled anchors are used mainly for memory update and local adaptation, while the label-free physical sensing route is preserved. This prevents a few labels from overriding target-domain sensing evidence. The feedback verifier enables local residuals, neighborhood propagation, or conservative fusion only when they improve feedback anchors without harming unlabeled stability.

\subsection{Verified Prediction Routes}

The tools above provide local evidence or candidate logits. SenseAgent produces final predictions only through a verified route, which is proposed by the planner and checked by the verifier. We summarize the three route types here to clarify the end-to-end data flow.

\textbf{Raw route.}
The raw route directly uses the frozen PatchTST. It is the conservative default when target memory indicates that raw inference is stable or when specialized actions fail verification.

\textbf{Gravity-aware route.}
The gravity-aware route uses gravity-derived evidence for prediction under orientation- and placement-related shifts. 
It first uses the gravity tool to construct target gravity views and reliability diagnostics. For low-motion windows, the route uses static geometry evidence to address lying-versus-upright ambiguity. For dynamic windows, it uses gravity-view embeddings and gravity-view prototype matching. The route-level verifier decides whether gravity-derived evidence is safe to use, and the implementation check selects the supported gravity-view encoder.

\textbf{Safe-groupwise route.}
The safe-groupwise route separates static and dynamic windows using the motion model and applies groupwise arbitration within each group.
Static windows may remain on raw inference or use conservative geometry- or style-based candidate logits. Dynamic windows may use raw logits, raw-prototype logits, gravity-view prototype logits, or style logits, depending on group utility and verifier support. 
The selected group logits are then merged into the final prediction.

These routes are not fixed dataset-level rules or manually selected adaptation modes. They are executable outcomes of the report-action-verification loop: the verified memory produces evidence, the planner proposes a route, the verifier checks no-harm conditions, and only the approved route can affect predictions.

\subsection{Scarce-feedback Adaptation}

SenseAgent also supports deployment with a small feedback budget $N$, which is the number of user-provided labeled data. 
Without feedback, the system follows the label-free pipeline: target windows update the online memory $\mathcal{M}_T$, the diagnosis report $R_T$ guides route proposal and tool selection, and the verifier accepts a final route. 
With feedback, the same pipeline is preserved. The labeled anchors serve only as sparse validation evidence rather than for backbone retraining or direct replacement of the label-free prediction route.

Given the anchors, SenseAgent evaluates verified routes and local correction candidates by their anchor accuracy and the log probability assigned to the anchor labels. 
This feedback evidence can guide memory actions, such as freezing the current memory, appending new target windows, or appending only reliable windows. A proposed memory state is accepted only when it preserves label-free stability and does not reduce feedback-anchor utility.


After the prediction route is verified, SenseAgent treats feedback adaptation as an optional residual correction on top of the verified label-free logits. It considers conservative local corrections around the labeled anchors, including local residual adjustment, same-motion-group propagation, tool-logit fusion, and fusion followed by propagation. 
These candidates operate on logits rather than model weights. A feedback verifier selects a candidate only if it improves anchor utility while preserving unlabeled stability. If no candidate passes this no-harm check, SenseAgent disables feedback adaptation and keeps the verified label-free logits.

\section{Experimental Settings}\label{sec:experiment}

All experiments are conducted on a Linux server with an Intel Xeon Gold 6258R CPU and NVIDIA A100 GPUs.
Each IMU window has $L=120$ time steps and six channels from tri-axial accelerometer and gyroscope readings. Since class distributions can be imbalanced across domains, we use macro-F1 as the evaluation metric.
SenseAgent uses PatchTST~\cite{nie2022time} as the raw IMU backbone and Qwen3.5-9B~\cite{yang2025qwen3,qwen35_9b} as the LLM planner, but the framework is backbone-flexible and LLM-agnostic: any temporal encoder and any instruction-following LLM can be substituted.


\begin{table}[t]
\centering
\caption{Dataset partitions used in our experiments. Each split entry reports the number of domains and windows.}
\resizebox{\columnwidth}{!}{
\begin{tabular}{lccccc}
\toprule
\textbf{Dataset} & \textbf{Classes} & \textbf{Source} & \textbf{User OOD} & \textbf{Pos./Dev. OOD} & \textbf{Compound OOD} \\
\midrule
HHAR & 6 & 25 / 13.0K & 14 / 7.2K & 21 / 13.9K & 7 / 4.3K \\
TNDA & 8 & 60 / 21.9K & 40 / 15.0K & 90 / 32.8K & 60 / 22.5K \\
PAMAP2 & 12 & 10 / 9.6K & 6 / 6.1K & 5 / 5.7K & 3 / 3.4K \\
Feng & 12 & 120 / 412.8K & 80 / 285.8K & 135 / 464.4K & 90 / 321.5K \\
\bottomrule
\end{tabular}}
\label{tab:datasets}
\end{table}

\begin{table*}[t]
\centering
\caption{Cross-domain macro-F1 comparison across four IMU sensing datasets. Dataset-specific columns report the mean across target domains. Avg. columns report macro-F1 averaged over all target domains in the corresponding OOD setting. The best result in each column is highlighted in bold.}
\resizebox{\textwidth}{!}{
\begin{tabular}{lccc| ccc| ccc| ccc| ccc}
\toprule
\multirow{2}{*}{\textbf{Method} $\downarrow$}
& \multicolumn{3}{c|}{\textbf{Feng}}
& \multicolumn{3}{c|}{\textbf{PAMAP2}}
& \multicolumn{3}{c|}{\textbf{TNDA}}
& \multicolumn{3}{c|}{\textbf{HHAR}}
& \multicolumn{3}{c}{\textbf{Average}} \\
\cline{2-16}
& User & Pos. & Comp.
& User & Pos. & Comp.
& User & Pos. & Comp.
& User & Dev. & Comp.
& User & Pos./Dev. & Comp. \\
\midrule

GRU
& 0.8041 & 0.7613 & 0.7203
& 0.4634 & 0.0999 & 0.1193
& \textbf{0.8262} & 0.3928 & 0.3571
& 0.5810 & 0.7543 & 0.5390
& \textbf{0.7735} & 0.6154 & 0.5649 \\

PatchTST
& 0.7751 & \textbf{0.8002} & 0.7327
& 0.4973 & 0.1531 & 0.1910
& 0.8219 & 0.4473 & 0.4392
& 0.7430 & 0.7416 & \textbf{0.7285}
& 0.7734 & 0.6559 & 0.6123 \\

T3A
& 0.6904 & 0.7177 & 0.6757
& 0.3954 & 0.1756 & 0.2478
& 0.7385 & 0.4397 & 0.4433
& 0.5159 & 0.5437 & 0.5220
& 0.6741 & 0.5927 & 0.5738 \\

SimCLR
& 0.7605 & 0.7829 & 0.7158
& 0.4450 & 0.1847 & 0.2100
& 0.7869 & 0.4100 & 0.3953
& 0.6601 & 0.6895 & 0.6933
& 0.7445 & 0.6295 & 0.5851 \\

CrossHAR-III
& 0.7487 & 0.7374 & 0.6780
& 0.3743 & 0.1745 & 0.1729
& 0.7337 & 0.2996 & 0.2801
& 0.6479 & 0.6350 & 0.6391
& 0.7183 & 0.5606 & 0.5176 \\

CrossHAR-II
& 0.7379 & 0.7564 & 0.6771
& 0.3757 & 0.1931 & 0.1709
& 0.7571 & 0.3200 & 0.2752
& 0.6812 & 0.6317 & 0.6415
& 0.7222 & 0.5783 & 0.5154 \\

LIMU-BERT
& \textbf{0.8187} & 0.7821 & \textbf{0.7337}
& 0.4380 & 0.0681 & 0.0633
& 0.7677 & 0.3293 & 0.3047
& 0.5942 & \textbf{0.7932} & 0.6147
& 0.7654 & 0.6064 & 0.5550 \\

\midrule

SenseAgent
& 0.7691 & 0.7998 & 0.7317
& \textbf{0.5110} & \textbf{0.2972} & \textbf{0.4089}
& 0.8188 & \textbf{0.5263} & \textbf{0.5518}
& \textbf{0.7432} & 0.7421 & \textbf{0.7285}
& 0.7697 & \textbf{0.6869} & \textbf{0.6581} \\

\bottomrule
\end{tabular}\label{tab:baseline_results}}
\end{table*}

\subsection{Datasets}


We evaluate SenseAgent on four public IMU activity recognition datasets with diverse activities, users, devices, and sensor placements, summarized in Table~\ref{tab:datasets}. 
For each dataset, we define domains by user identity and sensing condition, where the latter denotes device family in HHAR and body position or placement in TNDA, Feng, and PAMAP2.
Therefore, we construct three deployment shifts: 
\textit{User OOD} evaluates reserved users under source sensing conditions. 
\textit{Position/Device OOD} evaluates source users under reserved sensing conditions.
\textit{Compound OOD} evaluates reserved users under reserved sensing conditions, where both factors are unseen during source training.
Each target domain is evaluated independently.


\textbf{HHAR.} 
The HHAR~\cite{hhar2015dataset} dataset contains data from 9 users performing 6 activities: biking, sitting, standing, walking, walking upstairs, and walking downstairs. The data were collected using heterogeneous Samsung and LG smartphones and smartwatches. 
We use users \texttt{a--f} with Samsung devices as source domains, and reserve users \texttt{g--i} and LG devices for OOD evaluation.

\textbf{TNDA.}
TNDA~\cite{tnda2021dataset} contains wearable IMU recordings for 8 activities: sitting, standing, lying, upstairs, downstairs, cycling, walking, and jogging. The processed files used in our experiments contain 50 subject identifiers. We use subjects \texttt{1--30} with right-wrist and right-arm sensors as source domains, reserving subjects \texttt{31--50} and three unseen positions: back, left ankle, and left knee. 

\textbf{PAMAP2.}
PAMAP2~\cite{pamap2012dataset} contains 12 activities recorded with wearable IMU sensors at multiple body locations. 
We use subjects \texttt{101--105} with hand and ankle positions as source domains, and reserve subjects \texttt{106--108} and the chest position for OOD evaluation. 

\textbf{Feng.}
The Feng dataset~\cite{Feng2025} contains 12 activities collected from 25 users across 17 body positions. 
We use users \texttt{P01--P15} with eight body positions as source domains, and reserve users \texttt{P16--P25} and the remaining nine positions for OOD evaluation. 
The source positions include upper back, lower back, left/right wrist, left thigh, left shank, left foot, and right shoulder. 
\section{Experimental Evaluation}\label{sec:evaluation}

\subsection{Zero Feedback Performance}


We compare SenseAgent with representative zero-feedback baselines, where no user-provided target labels are available ($N=0$), although methods may observe the unlabeled target stream as defined in Sec.~\ref{sec:problem}.
Source-only supervised baselines include GRU~\cite{cho2014learning} and PatchTST~\cite{nie2022time}. 
Self-supervised baselines include SimCLR~\cite{chen2020simple}, LIMU-BERT~\cite{xu2021limu}, and CrossHAR~\cite{hong2024crosshar}. For fairness, SimCLR uses the same PatchTST backbone as SenseAgent. 
For CrossHAR, we report both source-only Paradigm III and target-unlabeled Paradigm II.
We also include T3A~\cite{iwasawa2021test}, a test-time adaptation baseline that updates the classifier using unlabeled target samples. 

Table~\ref{tab:baseline_results} reports the overall results.
Both supervised and self-supervised models can perform well in specific settings, such as LIMU-BERT on Feng user OOD and HHAR device OOD, or GRU on TNDA user OOD. 
However, their performance does not transfer uniformly to other shifts. For example, LIMU-BERT drops substantially on PAMAP2 position and compound OOD, while CrossHAR is below the raw PatchTST backbone in most settings. This supports our motivation that cross-domain IMU difficulty is target-dependent and cannot be addressed by selecting one fixed representation or adaptation rule in advance.

Unlabeled target data alone is also insufficient. For instance, T3A and CrossHAR-II both leverage information from unlabeled target data, but their average macro-F1 is lower than the raw PatchTST. This suggests that target-side adaptation can introduce negative transfer when the target evidence does not match the assumptions of the adaptation rule. SenseAgent also observes the unlabeled target stream, but uses it to build target memory and diagnostic reports before changing the prediction route. Its planner and verifier only accept an adaptation action when the evidence supports it, which helps preserve strong raw performance on easier shifts, such as Feng and HHAR, while enabling adaptation on harder shifts.

SenseAgent brings the largest gains under severe placement and compound shifts. 
On PAMAP2, it improves over PatchTST from 0.1531 to 0.2972 on position OOD and from 0.1910 to 0.4089 on compound OOD. On TNDA, it improves from 0.4473 to 0.5263 on position OOD and from 0.4392 to 0.5518 on compound OOD. These results suggest that placement and compound shifts are harder for a single fixed model and benefit more from SenseAgent's selective adaptation.
Overall, the results show that SenseAgent improves cross-domain IMU sensing not by simply using target data, but by using it selectively and safely.

\begin{table*}[t]
\centering
\caption{Scarce-feedback macro-F1 across OOD target domains. Rows with $N=1$ or $N=8$ use randomly sampled labeled anchors per target domain. $\dagger$ PTN-1shot uses one labeled support example per class and is therefore a stronger class-balanced few-shot reference. The best result in each column is highlighted in bold and the second-best result is underlined.}
\resizebox{\textwidth}{!}{
\begin{tabular}{lccc| ccc| ccc| ccc| ccc}
\toprule
\multirow{2}{*}{\textbf{Method} $\downarrow$}
& \multicolumn{3}{c|}{\textbf{Feng}}
& \multicolumn{3}{c|}{\textbf{PAMAP2}}
& \multicolumn{3}{c|}{\textbf{TNDA}}
& \multicolumn{3}{c|}{\textbf{HHAR}}
& \multicolumn{3}{c}{\textbf{Average}} \\
\cline{2-16}
& User & Pos. & Comp. & User & Pos. & Comp. & User & Pos. & Comp. & User & Dev. & Comp. & User & Pos./Dev. & Comp. \\
\midrule

T3A+FB ($N=1$)
& 0.6872 & 0.7162 & 0.6736
& 0.3956 & 0.1876 & 0.2517
& 0.7429 & 0.4630 & 0.4684
& 0.5191 & 0.5490 & 0.5335
& 0.6738 & 0.6009 & 0.5826 \\

T3A+FB ($N=8$)
& 0.6866 & 0.7149 & 0.6770
& 0.3819 & 0.2837 & 0.3138
& 0.7710 & 0.5777 & 0.5793
& 0.5638 & 0.5930 & 0.5805
& 0.6854 & 0.6469 & 0.6293 \\

Linear Probe ($N=8$)
& 0.2709 & 0.3013 & 0.2651
& 0.1665 & 0.2204 & 0.1898
& 0.4134 & 0.4067 & 0.4050
& 0.4245 & 0.4606 & 0.4613
& 0.3225 & 0.3508 & 0.3247 \\

PTN 1-shot$^\dagger$
& 0.6374 & 0.6902 & 0.6102
& 0.3704 & \textbf{0.4669} & \underline{0.4359}
& 0.7207 & \textbf{0.6926} & \textbf{0.7107}
& 0.5556 & 0.5778 & 0.6668
& 0.6416 & 0.6772 & 0.6471 \\
\midrule

SenseAgent ($N=1$)
& \underline{0.7655} & \underline{0.8016} & \underline{0.7263}
& \underline{0.5276} & 0.3128 & 0.4031
& \underline{0.8184} & 0.5397 & 0.5613
& \underline{0.7447} & \underline{0.7451} & \underline{0.7296}
& \underline{0.7683} & \underline{0.6932} & \underline{0.6585} \\

SenseAgent ($N=8$)
& \textbf{0.7678} & \textbf{0.8072} & \textbf{0.7347}
& \textbf{0.5487} & \underline{0.3738} & \textbf{0.4402}
& \textbf{0.8312} & \underline{0.5927} & \underline{0.6156}
& \textbf{0.7687} & \textbf{0.7579} & \textbf{0.7454}
& \textbf{0.7766} & \textbf{0.7175} & \textbf{0.6850} \\

\bottomrule
\end{tabular}\label{tab:scarce_feedback_results}}
\end{table*}

\begin{figure}[h]
    \centering
    \begin{subfigure}{0.48\columnwidth}
        \centering
        \includegraphics[width=\linewidth]{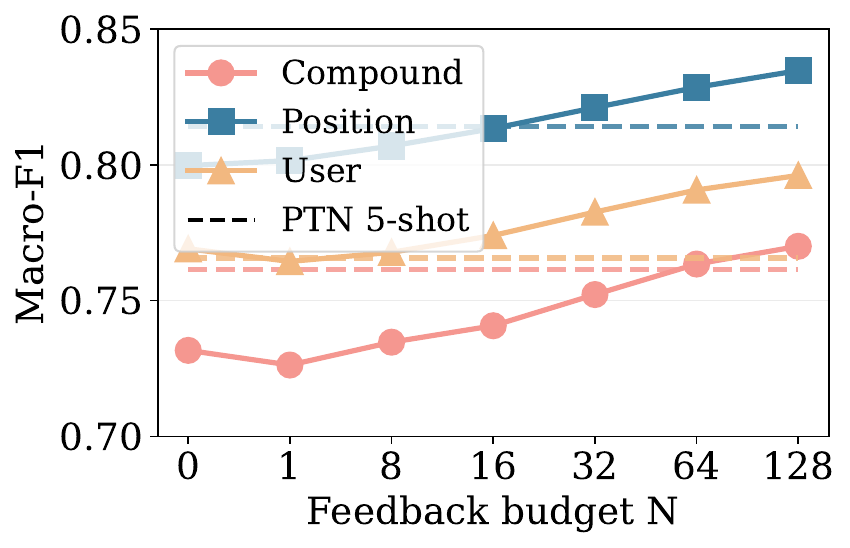}
        \caption{Feng}
        \label{fig:nsweep_feng}
    \end{subfigure}
    \hfill
    \begin{subfigure}{0.48\columnwidth}
        \centering
        \includegraphics[width=\linewidth]{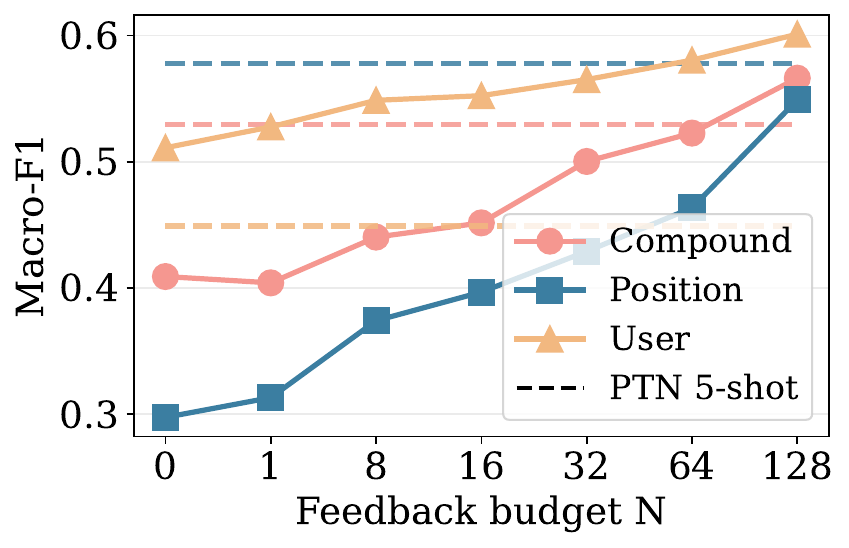}
        \caption{PAMAP2}
        \label{fig:nsweep_pamap2}
    \end{subfigure}

    \begin{subfigure}{0.48\columnwidth}
        \centering
        \includegraphics[width=\linewidth]{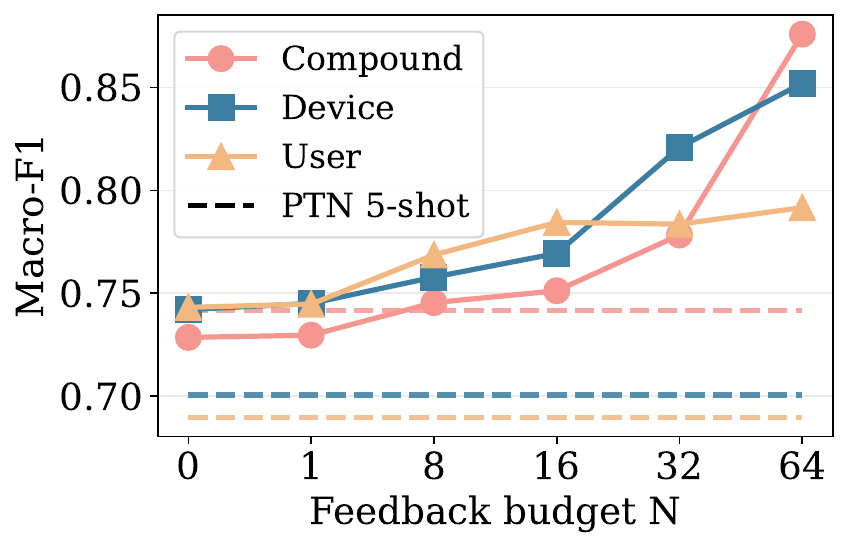}
        \caption{HHAR}
        \label{fig:nsweep_hhar}
    \end{subfigure}
    \hfill
    \begin{subfigure}{0.48\columnwidth}
        \centering
        \includegraphics[width=\linewidth]{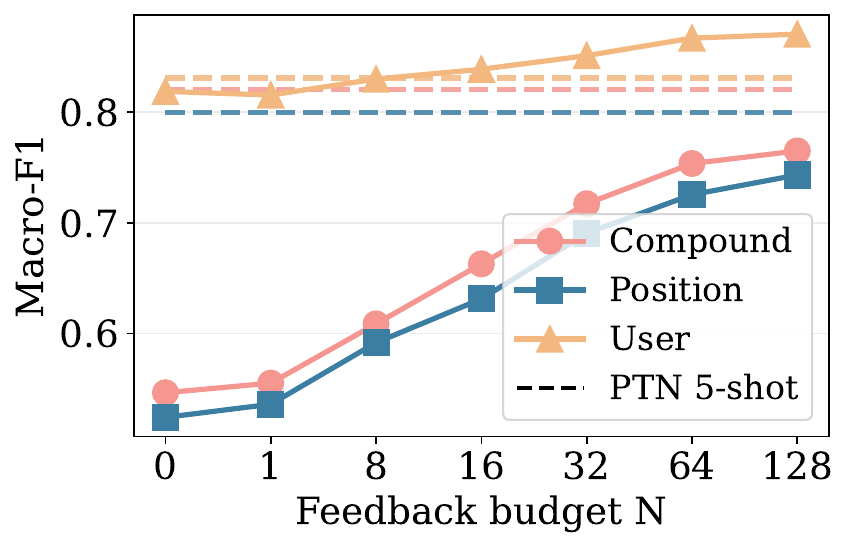}
        \caption{TNDA}
        \label{fig:nsweep_tnda}
    \end{subfigure}
    \caption{Effect of feedback budget on OOD adaptation. Solid lines show SenseAgent; dashed lines show PTN 5-shot. HHAR is evaluated up to $N=64$ due to limited target samples.}
    \label{fig:nsweep}
\end{figure}

\subsection{Scarce User Feedback Adaptation}

We evaluate scarce feedback by randomly sampling $N$ labeled anchors from each target domain, with all remaining target samples held out for evaluation.
We use $N=1$ and $N=8$ in this section. Since the datasets contain 6, 8, 12, and 12 activity classes, $N=8$ gives a budget close to one example per class on average, but unlike class-balanced few-shot support sets, random anchors do not guarantee class coverage. We repeat the sampling three times and report the average macro-F1.

We compare SenseAgent with three target-adaptation baselines. 
The prototypical network (PTN)~\cite{snell2017prototypical} is a classic few-shot learning method that constructs class prototypes from labeled target support examples. We note that its standard 1-shot setting follows a stronger label protocol than ours, since it provides one labeled example per class rather than one randomly sampled label per domain. 
Frozen linear probing trains only a linear classifier on top of frozen PatchTST embeddings using the sampled anchors, without updating the encoder. We omit its $N=1$ result because a single random anchor cannot train a multi-class linear head.
T3A+feedback extends test-time classifier adjustment~\cite{iwasawa2021test} by first using unlabeled target windows to adjust class prototypes and then using the labeled anchors to correct the prototypes of observed classes.

\textbf{Comparison under scarce feedback.}
Table~\ref{tab:scarce_feedback_results} reports the results for $N=1$ and $N=8$. 
PTN 1-shot is strong on several position and compound OOD scenarios, especially on PAMAP2 and TNDA. This is expected because its support set is class-balanced by construction. However, it is less effective on several Feng and HHAR domains, showing that class-balanced few-shot prototypes do not provide a uniformly strong solution across datasets.
T3A+feedback improves from $N=1$ to $N=8$ on several settings, especially on TNDA and HHAR, showing that additional labeled anchors can provide modest gains. 
However, direct classifier retraining remains unstable. With $N=8$, frozen linear probing still observes only a subset of classes in many target domains, and the resulting classifier can overfit the observed classes while ignoring unseen ones. This shows that scarce feedback can reduce target uncertainty, but an unsafe update rule can still cause negative transfer.

SenseAgent leverages labeled feedback more selectively. Rather than retraining the classifier directly, it stores anchors in target memory, applies local corrections only when supported, and verifies whether a feedback-based action should affect the prediction route. As a result, SenseAgent achieves the best average macro-F1 under the random-anchor protocol. With $N=8$, it obtains the highest average performance on user OOD, position/device OOD, and compound OOD. Even with only $N=1$, SenseAgent remains competitive with or better than the stronger PTN 1-shot reference on average position/device and compound shifts.

\textbf{Effect of feedback budget.}
To evaluate how SenseAgent benefits from different levels of user interaction, we vary the feedback budget $N$ in Fig.~\ref{fig:nsweep}.
Across datasets, increasing $N$ generally improves macro-F1, showing that SenseAgent can incorporate additional anchors without destabilizing the prediction route. 
The gains are more pronounced on PAMAP2 and TNDA, where low-feedback predictions leave a larger room for target-specific correction. HHAR benefits substantially from larger budgets on compound and device shifts. Feng exhibits a more gradual trend, suggesting that the initial route is already relatively reliable and additional anchors mainly provide incremental refinement.

Compared with the PTN 5-shot reference, SenseAgent remains competitive despite using randomly sampled rather than class-balanced anchors. 
It surpasses PTN with moderate budgets in several TNDA and HHAR settings, but remains weaker on TNDA position and compound shifts. 
This may be because these shifts strongly alter class-level motion prototypes, and PTN receives 5 target examples per class to directly build class-balanced prototypes. 
SenseAgent uses randomly sampled anchors, which can be imbalanced across activities, so its verifier applies feedback updates conservatively to avoid over-correction under uneven support.

Overall, the results show that scarce feedback is useful, but only when integrated selectively and safely. SenseAgent benefits from user anchors without directly overfitting to the limited observed classes, supporting the need for verifier-controlled memory and no-harm feedback integration.


\subsection{Runtime Decision Analysis}

We further examine whether SenseAgent adapts its runtime behavior to different target shifts, rather than executing a fixed adaptation route. Table~\ref{tab:runtime_decisions} summarizes the final verified routes and no-harm diagnostics under the zero feedback case.
The route column shows the final route after verification: Raw directly uses the frozen PatchTST prediction, Safe applies conservative static/dynamic groupwise arbitration, and Gravity applies the gravity-aware prototype route. Verifier override measures how often the verifier rejects a planner-selected gravity action and redirects it to a safer route before prediction. Gravity harm and final no-harm are post-hoc diagnostics computed with target labels only after evaluation. Gravity harm counts runs where forcing the Gravity route would perform worse than raw PatchTST, while final no-harm counts runs where the final SenseAgent output is no worse than raw PatchTST.

\begin{table}[t]
\centering
\caption{Runtime route selection and no-harm diagnostics under zero feedback. Verifier override measures interventions on unsupported gravity actions. Gravity harm and final no-harm are post-hoc evaluation diagnostics.}
\resizebox{\columnwidth}{!}{
\begin{tabular}{lcccc}
\toprule
\makecell{\textbf{Dataset}\\}
& \makecell{\textbf{Final Route (\%)}\\\textbf{Raw / Safe / Grav.}}
& \makecell{\textbf{Verifier}\\\textbf{Override (\%)}}
& \makecell{\textbf{Gravity}\\\textbf{Harm (\%)}}
& \makecell{\textbf{Final}\\\textbf{No-Harm (\%)}} \\
\midrule
Feng   & 0.5 / 97.8 / 1.6  & 93.7 & 98.1 & 93.2 \\
HHAR   & 28.6 / 71.4 / 0.0 & 71.4 & 61.9 & 100.0 \\
PAMAP2 & 0.0 / 21.4 / 78.6 & 0.0  & 7.1  & 92.9 \\
TNDA   & 0.0 / 8.8 / 91.2  & 4.2  & 28.6 & 75.3 \\
\bottomrule
\end{tabular}}
\label{tab:runtime_decisions}
\end{table}

The route distribution varies substantially across datasets, showing that SenseAgent is not following a fixed adaptation preference. 
Gravity route is frequently retained on PAMAP2 and TNDA, accounting for 78.6\% and 91.2\% of runs, respectively. In contrast, SenseAgent rarely keeps the gravity route on Feng and never keeps it on HHAR. This contrast shows that SenseAgent does not trigger gravity-aware adaptation solely based on the existence of position shift. Feng is an informative case: although it includes position OOD domains, its source domains already cover a broad set of body positions. This gives the raw backbone broader exposure to position variation during source training. Consistent with this, forcing gravity is harmful in 98.1\% of Feng runs, and the verifier overrides unsupported gravity actions in 93.7\% of runs. HHAR also shows high gravity harm at 61.9\%, suggesting that its target shifts are not reliably corrected by gravity-aware routing.

These results highlight the role of verifier-controlled execution. The planner may propose the gravity route when target diagnostics make it plausible, but the verifier decides whether the observed target evidence is sufficient before the route affects predictions. 
Overall, this analysis shows that SenseAgent benefits from selective execution rather than aggressive adaptation: fixed gravity routing would be unsafe on several datasets, while always suppressing gravity would miss useful adaptation opportunities.

\subsection{LLM Planner Analysis}\label{sec:llm_analysis}

We next analyze the role of the LLM planner. Since SenseAgent uses the LLM only as a structured route planner rather than a classifier, we ask two questions: (i) whether an LLM planner is more effective than a deterministic router given the same tools and verifier, and (ii) whether stronger LLMs further improve performance.
We also include a non-deployable oracle that uses target labels only after evaluation to choose the best executable route for each target domain, providing an upper bound on route selection.

\begin{table}[th]
\centering
\caption{No-LLM planner baseline under zero feedback. Values are macro-F1 differences relative to raw model. Prompt-Rule uses the same reports, tools, memory, and verifier as SenseAgent but replaces the LLM planner with deterministic rules. Oracle is a post-hoc upper bound.}
\resizebox{\columnwidth}{!}{
\begin{tabular}{lccccc}
\toprule
\textbf{Router} & \textbf{Feng} & \textbf{HHAR} & \textbf{PAMAP2} & \textbf{TNDA} & \textbf{Overall} \\
\midrule
Prompt-Rule & +0.0017 & +0.0002 & +0.0291 & +0.0051 & +0.0034 \\
SenseAgent & -0.0021 & +0.0003 & +0.1040 & +0.0724 & +0.0265 \\
Oracle & +0.0029 & +0.0650 & +0.1203 & +0.0919 & +0.0413 \\
\bottomrule
\end{tabular}}
\label{tab:planner_baseline}
\end{table}

\textbf{LLM vs. Rule-based.}
Table~\ref{tab:planner_baseline} reports gains over Raw. The deterministic Prompt-Rule router is conservative: it stays close to Raw on Feng and HHAR and yields limited gains on PAMAP2 and TNDA. In contrast, SenseAgent obtains much larger gains on the harder PAMAP2 and TNDA shifts, narrowing the gap to the oracle selector. This suggests that the LLM planner is useful when multiple label-free diagnostics conflict and a fixed rule underuses beneficial adaptation routes.

\begin{table}[th]
\centering
\caption{Difference between GPT-5.4 and Qwen3.5-9B planners on TNDA and HHAR. Values are weighted by the number of domains. Positive values indicate GPT-5.4 is higher.}
\resizebox{\columnwidth}{!}{
\begin{tabular}{lcccc}
\toprule
\textbf{Dataset} & \textbf{$N$} & \textbf{$\Delta$ SenseAgent} & \textbf{$\Delta$ Feedback-Adaptive} & \textbf{$\Delta$ Verifier Override} \\
\midrule
\multirow{3}{*}{TNDA}
& 0 & +0.0027 & +0.0027 & -0.0298 \\
& 1 & +0.0035 & +0.0034 & -0.0544 \\
& 8 & +0.0003 & +0.0009 & -0.0526 \\
\midrule
\multirow{3}{*}{HHAR}
& 0 & -0.0002 & -0.0002 & -0.4524 \\
& 1 & 0.0000 & 0.0000 & -0.5079 \\
& 8 & -0.0002 & +0.0001 & -0.5794 \\
\bottomrule
\end{tabular}}
\label{tab:stronger_planner_delta}
\end{table}

\textbf{Planner capacity and verification.}
Table~\ref{tab:stronger_planner_delta} shows that replacing Qwen3.5-9B with GPT-5.4~\cite{openai2026gpt54} changes final macro-F1 only marginally, but consistently reduces verifier overrides, especially on HHAR. This indicates that once the planner can produce valid structured actions, verifier-bounded tool execution largely governs final predictions, while stronger planners mainly improve proposal-verifier alignment. Together, the two tables support the role of the LLM agent as a flexible report-to-action planner, with sensing tools providing grounded computation and verifiers preventing unsupported adaptation from causing negative transfer.

\subsection{Ablation Study}

\begin{figure}[t]
    \centering
    \begin{subfigure}{0.49\columnwidth}
        \centering
        \includegraphics[width=\linewidth]{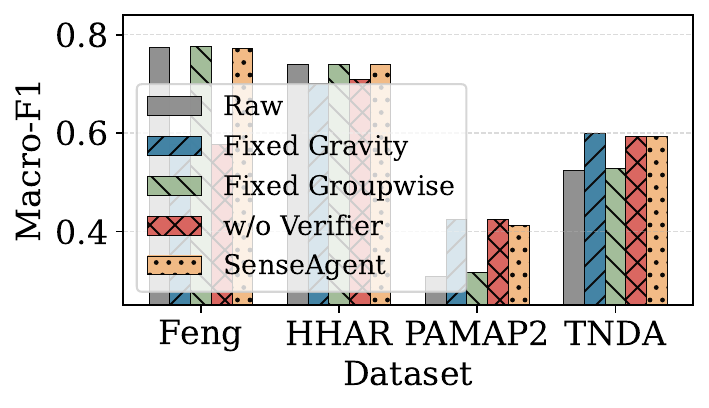}
        \caption{N=0}
        \label{fig:route_ablation_0}
    \end{subfigure}
    \hfill
    \begin{subfigure}{0.49\columnwidth}
        \centering
        \includegraphics[width=\linewidth]{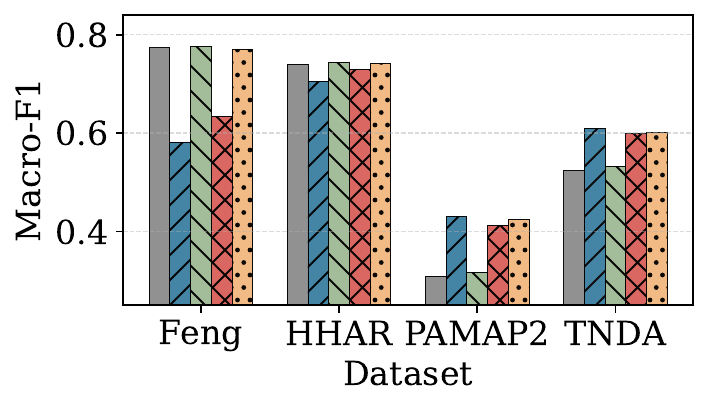}
        \caption{N=1}
        \label{fig:route_ablation_1}
    \end{subfigure}
    \caption{Route ablation under zero-feedback and scarce-feedback settings. SenseAgent uses verified routing to keep the beneficial route and suppress harmful adaptation.}
    \label{fig:ablation}
\end{figure}

In this section, we study two questions: whether SenseAgent needs runtime route selection, and which components contribute to the verified route. 
For route-level ablation, we compare SenseAgent with four variants: Raw always uses the frozen PatchTST, Fixed Gravity always uses the gravity-aware route, Fixed Groupwise applies static/dynamic groupwise arbitration, and w/o Verifier keeps the route without the no-harm check from verifier.

\textbf{Route-level Analysis.}
Fig.~\ref{fig:ablation} shows that no fixed route is consistently safe. 
Fixed Gravity improves PAMAP2 and TNDA, reaching 0.425 and 0.598 macro-F1 under $N=0$, but degrades Feng and HHAR to 0.576 and 0.702. Fixed Groupwise better preserves Feng and HHAR, but loses much of the gravity-related gain on PAMAP2 and TNDA. Raw PatchTST shows the opposite pattern: it remains strong on Feng and HHAR, but underperforms on the PAMAP2 and TNDA shifts.
These results confirm that target-domain shifts require different sensing actions, and applying one adaptation rule everywhere can cause negative transfer.
The route-level verifier addresses this issue by checking planner-proposed routes before execution.
Without it, performance drops toward harmful gravity behavior on Feng, from 0.772 to 0.576 under $N=0$ and from 0.770 to 0.634 under $N=1$. Meanwhile, SenseAgent still preserves most gravity benefits on PAMAP2 and TNDA, achieving 0.413 and 0.593 under $N=0$, and 0.424 and 0.601 under $N=1$.



\textbf{Component-level Analysis.}
Table~\ref{tab:component_ablation} shows that component contributions are shift-dependent.
Gravity-view components are most important on PAMAP2 and TNDA, where removing the LoRA adapter, gravity prototypes, or static geometry consistently reduces macro-F1.
This supports using gravity as a verified tool for orientation-aware shifts, rather than as a universal replacement for raw inference.
The gravity verifier has the largest effect on Feng and HHAR: removing it reduces Feng by 0.190 under $N=0$ and 0.132 under $N=1$, showing that verification is essential for avoiding negative gravity adaptation. Appendix~\ref{appx:threshold_sensitivity} further examines the sensitivity of the route-verifier thresholds.
Target memory updates mainly help PAMAP2 and TNDA, where target-side context is needed to estimate reliable gravity and prototype evidence. 
Source reliability has smaller and mixed effects, suggesting that it serves as a calibration signal rather than a standalone performance driver.

Overall, the ablation results show that SenseAgent's gains come from coordinating route selection, target memory, gravity evidence, and verification, rather than from any single component being universally best.

\begin{table}[t]
\centering
\caption{Component ablation results. Values are macro-F1 changes relative to the full SenseAgent, where negative values indicate removing the component hurts performance.}
\resizebox{\columnwidth}{!}{
\begin{tabular}{lcrrrrrr}
\toprule
\textbf{Data} & \textbf{N} & \makecell{\textbf{Gravity}\\\textbf{LoRA}} & \makecell{\textbf{Gravity}\\\textbf{Proto.}} & \makecell{\textbf{Gravity}\\\textbf{Verifier}} & \makecell{\textbf{Target}\\\textbf{Memory}} & \makecell{\textbf{Source}\\\textbf{Reliab.}} & \makecell{\textbf{Static}\\\textbf{Geometry}} \\
\midrule
Feng & 0 & 0.000 & +0.002 & -0.190 & -0.022 & +0.002 & +0.004 \\
HHAR & 0 & 0.000 & 0.000 & -0.031 & +0.006 & +0.001 & 0.000 \\
PAMAP2 & 0 & -0.107 & -0.104 & -0.071 & -0.046 & -0.016 & -0.051 \\
TNDA & 0 & +0.004 & -0.074 & -0.008 & -0.056 & +0.018 & -0.042 \\
\midrule
Feng & 1 & 0.000 & +0.005 & -0.132 & -0.014 & +0.001 & +0.005 \\
HHAR & 1 & 0.000 & 0.000 & -0.013 & +0.005 & 0.000 & 0.000 \\
PAMAP2 & 1 & -0.110 & -0.114 & -0.077 & -0.054 & -0.024 & -0.047 \\
TNDA & 1 & +0.006 & -0.078 & -0.007 & -0.058 & +0.016 & -0.033 \\
\bottomrule
\end{tabular}}
\label{tab:component_ablation}
\end{table}

\begin{table}[t]
\centering
\caption{Deployment overhead under zero feedback. Report E2E is measured per target report; cold LLM cost is measured per generated action.}
\resizebox{\columnwidth}{!}{
\begin{tabular}{ccccc}
\toprule
\makecell{\textbf{Raw Route}\\\textbf{/ Window}} &
\makecell{\textbf{Gravity Route}\\\textbf{/ Window}} &
\makecell{\textbf{Report}\\\textbf{E2E}} &
\makecell{\textbf{Cold LLM}\\\textbf{Generation}} &
\makecell{\textbf{In/Out}\\\textbf{Tokens}} \\
\midrule
7.82 ms & 14.91 ms & 1.76 s & 10.52 s & 2.87K / 60.3\\
\bottomrule
\end{tabular}}
\label{tab:deployment_overhead}
\end{table}

\subsection{Deployment Overhead}

In this section, we evaluate online deployment overhead under zero feedback, excluding offline source construction.
The online pipeline contains two levels of computation.
The streaming path performs window-level inference, while the report-level path performs diagnosis construction, tool execution, target-memory maintenance, verification, and final logit selection.

Table~\ref{tab:deployment_overhead} shows that SenseAgent keeps online classification lightweight.
Window-level inference takes 7.82~ms for the raw route and 14.91~ms for the gravity route. The per-window latency of the safe-groupwise route lies between the raw and gravity routes, depending on the selected tools. The gravity route is slower because it first constructs gravity-view channels and then runs the gravity-view encoder.
At the report level, a cold LLM planner call takes 10.52~s for a newly generated action with 2.87K input tokens and 60.3 output tokens on average.
In deployment, many IMU streams exhibit recurring diagnosis patterns across reports, allowing the system to reuse cached outputs. As a result, the report-level overhead is reduced to 1.76~s in our evaluation.
Overall, these results indicate that SenseAgent achieves acceptable deployment latency as an online adaptation system.

\section{Related Work}\label{sec:related}

\subsection{Deployment Heterogeneity in IMU Sensing}


IMU sensing remains highly sensitive to deployment conditions. Sensor readings may vary significantly between devices, body locations, sensor orientations, and user motion styles. Stisen et al. showed that heterogeneity between devices, operating systems, and workloads can significantly reduce the accuracy of activity recognition in large-scale deployments~\cite{stisen2015smart}. SenseHAR addressed variations in available devices, body locations, and sampling rates by mapping device-specific inputs into a shared virtual activity sensor representation~\cite{jeyakumar2019sensehar}. More recently, UniHAR further studied practical HAR adoption and introduced physics-informed data augmentation to improve generalization across users and datasets~\cite{xu2023practically}. 


\subsection{Cross-domain IMU Representation Learning}

Because it is expensive to collect and annotate labeled IMU data, self-supervised and cross-domain representation learning played important roles in wearable sensing. 
LIMU-BERT adapts the masked modeling idea from BERT to IMU streams and learns generalized representations from unlabeled sensor data for downstream IMU sensing tasks~\cite{xu2021limu}. ColloSSL leverages synchronized data from multiple devices worn by the same user to construct collaborative self-supervised signals~\cite{jain2022collossl}. Haresamudram et al. systematically evaluate self-supervised HAR methods and show that robustness must be assessed across source and target conditions, dataset characteristics, and feature-space behavior~\cite{haresamudram2022assessing}. CrossHAR further targets cross-dataset HAR by using hierarchical self-supervised pretraining to improve generalization to unseen target datasets~\cite{hong2024crosshar}. 

\begin{table}[th]
\centering
\caption{Comparison with LLM-based sensing systems.}
\label{tab:comparison}
\scriptsize
\begin{tabularx}{\columnwidth}{@{}>{\raggedright\arraybackslash}Xccccc@{}}
\toprule
\multirow{2}{*}{\textbf{Methods}} &
\textbf{Sensor} &
\textbf{LLM} &
\textbf{Runtime} &
\textbf{Target} &
\textbf{Adapt} \\
&
\textbf{Evidence} &
\textbf{Reason} &
\textbf{Plan} &
\textbf{Memory} &
\textbf{Verifier} \\
\midrule

Semantic HAR Alignment~\cite{yan2025large} &
\cmark &
\cmark &
\xmark &
\xmark &
\xmark \\

LLM-IMU Sensing~\cite{li2025sensorllm,hong2025llm4har,zhang2026sensorlm} &
\cmark &
\cmark &
\xmark &
\xmark &
\xmark \\

IMU Reasoning~\cite{asif2025llasa,zhang2025large} &
\cmark &
\cmark &
\xmark &
\xmark &
\xmark \\

IoT-LLM~\cite{an2026iot} &
\cmark &
\cmark &
\xmark &
\xmark &
\xmark \\

AutoIoT~\cite{shen2025autoiot} &
\xmark &
\cmark &
\cmark &
\xmark &
\xmark \\

\textbf{SenseAgent} &
\cmark &
\cmark &
\cmark &
\cmark &
\cmark \\

\bottomrule
\vspace{-15mm}
\end{tabularx}
\end{table}

\subsection{Semantic and LLM-based Sensing}

Recent sensing systems increasingly move beyond single-task classifiers toward foundation-style, multimodal, and semantic sensing. 
Babel aligns multiple sensing modalities, showing that aligned representations can improve both single-modality sensing and multimodal fusion~\cite{dai2025babel}. Related cross-modal systems in mobile and embedded sensing use one modality to improve another, such as IMU-to-Doppler adaptation for Doppler-based activity recognition~\cite{bhalla2021imu2doppler}. 
In parallel, language-guided HAR connects sensor readings and activity labels through semantic descriptions.
LanHAR uses LLM-generated semantic interpretations to mitigate cross-dataset heterogeneity and recognize new activities~\cite{yan2025large}. 
SensorLLM aligns motion sensor time series with natural-language trend descriptions for HAR~\cite{li2025sensorllm}, while LLM4HAR adapts pretrained LLMs for cross-domain, on-device HAR~\cite{hong2025llm4har}. LLaSA supports open-ended IMU reasoning and question answering~\cite{asif2025llasa}, and Zhang et al. use an LLM-driven agent network to interpret worker activities from a single head-mounted IMU~\cite{zhang2025large}. Beyond HAR-specific systems, recent LLM-based IoT work explores how LLMs can reason over real-world sensor data or generate AIoT applications from natural-language requirements~\cite{an2026iot,shen2025autoiot}. 


Overall, most existing methods either pursue robustness before deployment or use LLMs mainly for alignment, classification, explanation, and reasoning. Similarly, LLM-based IoT systems focus on sensor-aware reasoning or AIoT application programming rather than deployment-time IMU adaptation. As summarized in Table~\ref{tab:comparison}, SenseAgent extends this view from representation alignment and sensor understanding to agentic operation. Rather than directly classifying raw IMU windows, SenseAgent uses the LLM as a runtime planner that reasons over structured target-domain evidence and coordinates tool selection, adaptation, memory update, and verification in a closed loop.

\section{Discussion and Future Work}\label{sec:future}



\subsection{Leveraging Stronger LLMs}

Section~\ref{sec:llm_analysis} shows both the benefits and limitations of the current planner design. SenseAgent outperforms a deterministic prompt-rule router on harder placement-heavy shifts, indicating that LLM-guided routing helps arbitrate conflicting label-free diagnostics. However, replacing Qwen3.5-9B with GPT-5.4 brings only marginal macro-F1 gains, even though it reduces verifier overrides, because the current planner is intentionally constrained to a fixed report-to-action interface. It can select and configure predefined sensing routes and tools, but cannot expand the sensing action space. 
A future direction is to combine more capable LLM planners with stronger sensing harnesses, allowing the planner to compose sensing primitives or generate executable adaptation code while preserving auditability. Such actions should still pass through structured reports, execution logs, and verifier gates before affecting predictions, expanding LLM capability without turning the planner into an unconstrained sensor-data predictor.

\subsection{Learning the Planner and Verifier}
SenseAgent currently relies on prompted planner actions and verifier rules that are designed from sensing priors and calibrated source-domain episodes.
This makes the system deployable without target labels, but the report-to-action policy and verifier thresholds are not directly optimized from deployment outcomes.
A natural future direction is to improve both components from logged diagnosis reports, planner actions, verifier decisions, sparse user feedback, and delayed target performance.
For the planner, reinforcement learning or contextual bandit learning can optimize which action to select under each report state, such as when to use gravity-view inference, prototype transfer, memory updates, or conservative fallback.
For the verifier, logged outcomes can be used to calibrate the risk score of each proposed action and reduce both false rejections and unsafe acceptances.

\section{Conclusion}\label{sec:conclusion}
This paper presents SenseAgent, an LLM-guided adaptive sensing framework for cross-domain IMU activity recognition. SenseAgent keeps raw sensing computation local, uses the LLM for structured planning over sensing tools, and relies on a verifier to guard against harmful adaptation. By integrating source experience memory, target memory, physics-guided gravity reasoning, prototype and style tools, and scarce-feedback adaptation, SenseAgent reframes cross-domain IMU sensing as a closed-loop agentic sensing problem rather than a fixed adaptation pipeline. 
Experiments across diverse user, position, device, and compound shifts show that SenseAgent improves cross-domain IMU-based activity recognition while preserving auditable decisions and practical deployment overhead. 
These results highlight the potential of verified LLM-guided planning as a foundation for robust and adaptive sensing systems under real-world distribution shifts.


\bibliographystyle{ACM-Reference-Format}
\bibliography{Ref}


\appendix

\section{Planner Prompt Template}\label{appx:prompt}

SenseAgent uses the LLM as a structured planner rather than as an IMU classifier.
Table~\ref{tab:prompt_template} summarizes the conceptual prompt template used by the planner. For implementation, the \texttt{route} is serialized as a set of tool actions plus a gravity variant, and \texttt{feedback\_use} is resolved by verifier-controlled feedback adaptation rules.

\begin{table}[h]
\centering
\caption{Prompt template for the SenseAgent planner.}
\scriptsize
\begin{tabularx}{\columnwidth}{p{0.23\columnwidth}X}
\toprule
\textbf{Component} & \textbf{Template content} \\
\midrule
Role & You are SenseAgent, an action planner for cross-domain IMU activity recognition. You do not classify raw IMU windows directly; you propose a prediction route, a feedback-use policy, and a memory update from a structured diagnosis report. \\
\midrule
Input constraints & You only see a JSON diagnosis report containing label-free target evidence, source-calibrated route/tool summaries, memory statistics, gravity/prototype/style reliability, and optional user-provided feedback anchors. Do not use dataset name, target domain identity, user identity, device identity, body position, OOD type, or hidden target labels. \\
\midrule
Action space & Choose a route instantiated by tool actions from \texttt{raw\_patchtst}, \texttt{safe\_groupwise}, \texttt{gravity\_proto\_tool}, \texttt{raw\_proto\_tool}, and \texttt{style\_normalization}, with gravity variant \texttt{auto}, \texttt{lora}, or \texttt{base}; choose feedback use from conservative freeze, local adjustment, propagation, fusion, or fusion-propagation; and choose memory update from \texttt{append}, \texttt{append\_filtered}, and \texttt{freeze}. \\
\midrule
Decision rule & Use candidate utilities, raw stability, shift diagnostics, gravity reliability, prototype support, style reliability, and feedback evidence as weak evidence rather than fixed rules. Prefer conservative actions when raw inference is stable or support for specialized routes is weak. Proposed actions are checked by downstream verifiers and converted into final executable routes only if accepted. \\
\midrule
Output format & Return JSON only: \texttt{\{"route":..., "feedback\_use":..., "memory\_update":..., "rationale":...\}}. \\
\bottomrule
\end{tabularx}
\label{tab:prompt_template}
\end{table}

\section{Verifier Threshold Sensitivity}\label{appx:threshold_sensitivity}

We evaluate the sensitivity of the label-free route verifier by replaying zero-feedback planner proposals over saved diagnosis reports and materialized tool outputs. Table~\ref{tab:threshold_sensitivity} sweeps the static-support threshold $\tau_s$ and the raw-stability threshold $\tau_r$ used to block unsupported gravity routing. Macro-F1 remains stable across the middle range of thresholds, while overly permissive static support ($\tau_s=0.35$) accepts more unsupported gravity actions and reduces the no-harm rate.

\begin{table}[h]
\centering
\caption{Zero-feedback sensitivity of the route verifier thresholds. No-Harm reports the fraction of target domains whose replayed route is no worse than raw PatchTST.}
\small
\begin{tabular}{ccccc}
\toprule
\textbf{$\tau_s$} & \textbf{$\tau_r$} & \textbf{Macro-F1} & \textbf{No-Harm} & \textbf{Override} \\
\midrule
0.35 & 0.65 & 0.6657 & 0.734 & 0.458 \\
0.35 & 0.75 & 0.6655 & 0.733 & 0.457 \\
0.35 & 0.85 & 0.6607 & 0.684 & 0.390 \\
0.45 & 0.65 & 0.6935 & 0.854 & 0.601 \\
0.45 & 0.75 & 0.6933 & 0.851 & 0.598 \\
0.45 & 0.85 & 0.6882 & 0.799 & 0.517 \\
0.55 & 0.65 & \textbf{0.6996} & 0.889 & 0.658 \\
0.55 & 0.75 & 0.6992 & 0.884 & 0.652 \\
0.55 & 0.85 & 0.6945 & 0.832 & 0.566 \\
0.65 & 0.65 & 0.6932 & \textbf{0.906} & 0.756 \\
0.65 & 0.75 & 0.6958 & 0.902 & 0.732 \\
0.65 & 0.85 & 0.6941 & 0.854 & 0.624 \\
\bottomrule
\end{tabular}
\label{tab:threshold_sensitivity}
\end{table}

The middle threshold range preserves similar macro-F1 and no-harm behavior, suggesting that SenseAgent is not highly sensitive to a single verifier hyperparameter setting.

\end{document}